\documentclass[runningheads]{llncs}
\usepackage{graphicx}
\usepackage{dsfont}
\usepackage{amsmath}
\usepackage{float}
\usepackage{mathtools}
\usepackage[T1]{fontenc}
\usepackage{amsmath}
\usepackage{caption}
\usepackage{subcaption}
\usepackage{textalpha}
\usepackage{dsfont}
\usepackage{amsmath}
\usepackage{float}
\usepackage{multirow}
\usepackage{booktabs} % For formal tables
\usepackage{hhline}
\usepackage{longtable}
\usepackage{caption}
\usepackage{xcolor,pifont}
\usepackage{multicol}
\usepackage{tabularx}
\usepackage{amssymb}% http://ctan.org/pkg/amssymb
\usepackage{pifont}% http://ctan.org/pkg/pifont
\usepackage{adjustbox}
\usepackage{lipsum}
\usepackage{tikz}
\usepackage{enumitem}
\usepackage{booktabs}
\usepackage{xcolor,pifont}
\usepackage{amssymb}% http://ctan.org/pkg/amssymb
\usepackage{pifont}% http://ctan.org/pkg/pifont
\newcommand{\xmark}{\ding{55}}%
\newcommand*\colourcheck[1]{%
  \expandafter\newcommand\csname #1check\endcsname{\textcolor{#1}{\ding{52}}}%
}

\usepackage[colorlinks=true, citecolor=citecolor, linkcolor=linkcolor]{hyperref}
\definecolor{citecolor}{HTML}{0071BC}
\definecolor{linkcolor}{HTML}{ED1C24}
\usepackage{verbatim}
\usepackage[capitalize]{cleveref}

\begin{document}
%

%\title{A Domain Adaptive Hybrid Approach for Document Layout Analysis in Document images}
\title{A Hybrid Approach for Document Layout Analysis in Document images}
%\titlerunning{Domain Adaptive Hybrid Approach for Document Layout Analysis}
\titlerunning{A Hybrid Approach for Document Layout Analysis}
%\author{%Paper ID: 25
\author{Tahira Shehzadi*\inst{1,2,3}\orcidID{0000-0002-7052-979X} \and
Didier Stricker\inst{1,2,3} \and
Muhammad Zeshan Afzal\inst{1,2,3}\orcidID{0000-0002-0536-6867}}
\authorrunning{T. Shehzadi et al.}
\institute{Department of Computer Science, Technical University of Kaiserslautern, Germany \and
Mindgarage, Technical University of Kaiserslautern, Germany \and
German Research Institute for Artificial Intelligence (DFKI), 67663 Kaiserslautern, Germany\\
\email{\{tahira.shehzadi@dfki.de\}}
}
\maketitle              % typeset the header of the contribution
\begin{abstract}
Document layout analysis involves understanding the arrangement of elements within a document. This paper navigates the complexities of understanding various elements within document images, such as text, images, tables, and headings. The approach employs an advanced Transformer-based object detection network as an innovative graphical page object detector for identifying tables, figures, and displayed elements. We introduce a query encoding mechanism to provide high-quality object queries for contrastive learning, enhancing efficiency in the decoder phase. We also present a hybrid matching scheme that integrates the decoder's original one-to-one matching strategy with the one-to-many matching strategy during the training phase. This approach aims to improve the model's accuracy and versatility in detecting various graphical elements on a page. Our experiments on PubLayNet, DocLayNet, and PubTables benchmarks show that our approach outperforms current state-of-the-art methods. It achieves an average precision of \textbf{97.3\%} on PubLayNet, \textbf{81.6\%} on DocLayNet, and \textbf{98.6\%} on PubTables, demonstrating its superior performance in layout analysis. These advancements not only enhance the conversion of document images into editable and accessible formats but also streamline information retrieval and data extraction processes. 
\end{abstract}

\section{Introduction}
\begin{figure}
\centering
\includegraphics[width=0.99\textwidth]{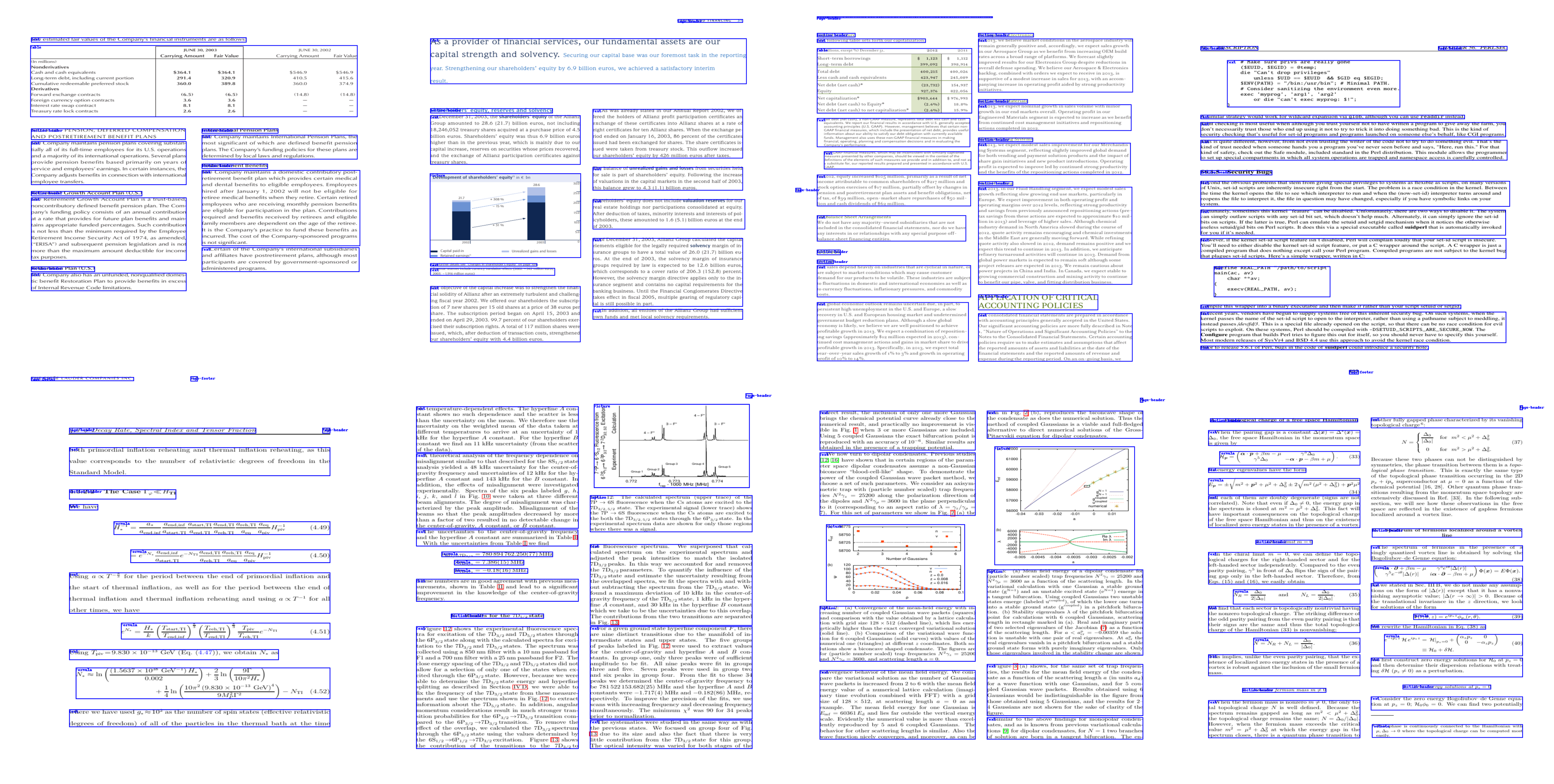}
\caption{Diverse layouts and element types in the DocLayNet Dataset, including elements such as captions, footnotes, formulas, and more. It underscores the challenges in document layout analysis, like interpreting dense text and categorizing diverse elements.}\label{fig:semi_gt}
\end{figure}
Systems for Document Intelligence (DI) is essential in enhancing the efficiency of automating large-scale document processing tasks, primarily focusing on extracting and understanding content within these documents. These systems are pivotal in key business intelligence operations such as document retrieval, text recognition, and content categorization, which rely heavily on extracting information and transforming documents into a structured, machine-readable format. This process seamlessly integrates the information extracted into further document processing workflows. As a result, significant improvements have been achieved across various industries, including banking, finance, and healthcare~\cite{AI_app,shehzadi_IEEE_I}. Document Layout Analysis (DLA) has become a key component in Document Intelligence due to its deriving structured formats from unstructured documents. This structuring is vital for accurately identifying and extracting essential document data. DLA encompasses two primary aspects: physical layout analysis, which identifies and spatially categorizes physical page elements like text, images, and tables, and logical layout analysis, which assigns semantic roles to these elements, such as titles, paragraphs, and headers, while understanding their hierarchical and reading order relationships. This analysis is essential for converting scanned documents into editable and searchable formats. However, it faces challenges due to the diversity of document layouts, the varying sizes and shapes of elements, and the complexity of accurately interpreting these elements across different documents.

Previously, remarkable progress has been made in document layout analysis through deep learning techniques, including advanced technologies like Faster RCNN~\cite{faster23} and Mask RCNN~\cite{mask86}, as well as other specialized frameworks~\cite{cascadercnn8,ShuffleNetv2}. These methods, effective in specific scenarios such as table detection and the layout analysis of academic papers~\cite{DeepDeSRT3,TDcl34,rethink78,semimask1,naik86}, sometimes face limitations in wider applications across various tasks~\cite{deep_review56}. The advancement of Transformer-based networks~\cite{deformable_detr,UPDETR_CVPR20,pnp6,yolos6,shehzadi_semi-detr_table,rego2,dino23,shehzadi_DT_review6} marks a significant advancement over traditional convolutional neural networks (CNNs), primarily due to their global attention mechanisms and Non-Maximum Suppression (NMS) free design. However, these models still show constraints in precisely detecting textual regions, especially in identifying small-scale text areas such as headers, footers, and section titles. For example, DINO~\cite{dino23}, a leading Transformer-based detection model~\cite{dino23}, experiences a notable drop in detection accuracy for these small text regions on the DocLayNet dataset~\cite{doclaynet_data}. Fig.~\ref{fig:semi_gt} shows complex layouts from DocLayNet, with details like captions and footnotes. To improve DLA, we need better algorithms for handling different documents, from academic papers to magazines.

In this paper, we propose an approach to address the challenges of document layout analysis, focusing on accurately identifying graphical elements within pages, such as tables, figures, and formulas. We employ an advanced Transformer-based object detection network~\cite{dino23}, for its exceptional capability in detecting various graphical page objects. Enhancing this capability, we introduce a Query Encoding Strategy to provide high-quality object queries by taking high-level query features from the backbone. These query features provide better predictions for small graphical objects like page headers, footers, and titles, combined with the decoder's original queries to improve overall performance. This mechanism is pivotal for contrastive learning, significantly improving the efficiency of the model's decoder phase and enabling more effective processing of complex document layouts. Furthermore, our approach introduces a novel hybrid matching scheme that merges the decoder's original one-to-one matching strategy with an auxiliary one-to-many matching strategy. This integration, implemented during the training phase, is key to boosting the model's accuracy and adaptability in recognizing diverse classes of graphical elements. By combining the transformer's object detection capabilities with our unique encoding query and selection strategies, our method sets a new benchmark in document layout analysis, significantly advancing the field's ability to accurately detect and interpret graphical elements within various documents.

\noindent We summarize the main contributions of this paper as follows:
\begin{itemize}
  \item[$\bullet$] We introduce a Transformer-based framework for document layout analysis, incorporating a ResNet-50 backbone. This framework is augmented with an enhanced query encoding mechanism and innovative query-selection strategies. By integrating these strategies, our approach sets a new standard in document layout analysis. This significant advancement contributes to accurately detecting graphical elements in various document types.
    \item[$\bullet$] We present a unique query selection scheme that blends the decoder's original one-to-one matching strategy with a one-to-many matching strategy. This integration, crucial during the training phase, significantly enhances the model's accuracy and adaptability in detecting and categorizing various graphical elements across different documents.
    \item[$\bullet$] We introduce an enhanced query encoding mechanism to improve the efficiency of the model's decoder phase and enable more effective processing of complex document layouts.
    \item[$\bullet$]  To validate the effectiveness of our approach, we conduct comprehensive evaluations on three distinct datasets: PubLayNet, DocLayNet, and Pubtables. These evaluations demonstrate the robustness and applicability of our proposed method across various documents and layout challenges.
\end{itemize}

\section{Related Work}
\label{sec:Literature-Review}
Layout analysis is crucial to extract data from digital documents effectively. It involves understanding the spatial arrangement and relationships between various elements like tables, text, figures, and titles. Before deep learning approaches~\cite{faster23,mask86,retinaNet68}, heuristic rule-based algorithms~\cite{historical_m4,Languag_SV7} were emplyed in layout analysis. However, with technological advancements, convolutional neural networks (CNNs) became the primary method, providing significant improvements. More recently, transformer-based architectures~\cite{deformable_detr,smca23,CondDE,WBdetr4,pnp6,yolos6,fpdetr,rego2,dino23} have emerged as the leading approach, show remarkable effectiveness in this domain. This section aims to offer an in-depth review of these cutting-edge techniques, exploring a variety of approaches in Document Layout Analysis (DLA).\\

\noindent\textbf{Heuristic Rule-Based DLA.} The document layout analysis using heuristic techniques is generally categorized into top-down, bottom-up, and hybrid approaches. Bottom-up methods~\cite{historical_m4,Languag_SV7} involve elementary processes such as clustering and combining pixels to form uniform regions for akin objects while segregating dissimilar ones. Conversely, top-down approaches~\cite{Text_graphic6,seg_page_images6} iteratively divide the document image into various regions until distinct areas encompassing similar objects are formed. While bottom-up strategies are capable of handling intricate layouts, they require significant computational resources. On the other hand, top-down methods are more efficient in terms of implementation speed but lack versatility, showing optimal performance only with certain document types. Hybrid methods~\cite{TD_handwritten9,TD_Multipag_pdf5} combine the strengths of both bottom-up and top-down techniques, achieving both rapid and effective outcomes. Before the advent of deep learning, these heuristic strategies were the leading methods for detecting tables in documents.\\

\noindent\textbf{Deep Learning-based DLA.} With the rise of deep learning approaches~\cite{DeepDeSRT3,TDcl34,semimask1}, Convolutional Neural Networks (CNNs) have performed better than traditional rule-based algorithms in document analysis~\cite{continuaLR45,Ksemi76,CasTab45,Hyb65,cas10,TSRkhurm4,Real_DICls4}. This development represents a significant improvement in the precision and efficiency of processing and understanding complex document layouts. Introducing Faster-RCNN~\cite{faster23} marks a significant advancement in document object detection, facilitating effective page segmentation~\cite{page_se56}. Subsequently, Mask-RCNN~\cite{mask86} set a new benchmark in layout segmentation, particularly for newspapers. RetinaNet~\cite{retinaNet68} further contributes to this evolution by focusing on keyword detection in document images, although its complexity limits its application to text region detection. For table detection and structural recognition, DeepDeSRT~\cite{DeepDeSRT3} introduces an innovative image transformation approach that discerns table features for input into a fully convolutional network employing skip pooling. The ICDAR2017 POD (Page Object Detection) benchmark, introduced by Saha et al.~\cite{GOjt3}, utilizes a transfer learning-based Faster-RCNN architecture to detect elements like mathematical equations, tables, and figures. To address cross-domain challenges in Document Object Detection, a new benchmark~\cite{cross_domain_od5} is established, focusing on domain adaptation strategies. More recently, a vision-based layout detection benchmark~\cite{vision_LD6} employs a recurrent convolutional neural network with a VoVNet-v2 backbone, generating synthetic PDF documents from the ICDAR-2013 and GROTOAP datasets to set new standards in scientific document analysis.\\

\noindent\textbf{Transformer-Based DLA}. 
Document layout analysis is rapidly advancing with Transformer architectures, known for their positional embedding and attention mechanisms. These methods are known for their unique features like positional embedding and attention mechanism~\cite{att75}. DiT~\cite{Lidit78} has set a new standard in classifying document images, layout analysis, and table detection, employing self-supervised training on extensive collections of unlabeled document images. However, its application is limited to smaller datasets like PRIMA. Li et al.~\cite{StrucTexTST56} develop a method that combines different types of data to understand structured text in documents. However, this method struggles with text that has similar meanings.
Furthermore, the TILT~\cite{going_DI7} model simultaneously processes textual, visual, and layout data through an encoder-decoder Transformer setup. Another implementation of a transformer encoder-decoder in~\cite{trans_DLU5} establishes a benchmark for the PubLayNet dataset~\cite{PubLayNet3}, integrating text data extracted via OCR. The LayoutLMv3~\cite{layoutMV3} model improves visual document understanding by jointly learning from text, layout, and visual elements. It performs better with large datasets but has limitations with smaller ones. Other recent models~\cite{docformer59,Donut64,unidoc89,layoutlm56} also adopt joint pre-training strategies for various tasks, including document visual question answering. The transformer-based architectures have emerged as the leading approaches, show remarkable effectiveness in object detection domain~\cite{deformable_detr,UPDETR_CVPR20,smca23,CondDE,shehzadi_semi-detr_table,WBdetr4,pnp6,yolos6,fpdetr,bridging_per8,rego2,dino23}. However, when employing DINO~\cite{dino23} or other Transformer-based networks in document layout analysis, there's a noted limitation in their performance with small graphical objects like page titles, headers, and footers. To improve this, We enhance the hybrid query mechanism and matching scheme. This strategy elevates our document layout analysis, allowing for more precise and flexible detection and interpretation of various document graphical elements.
\section{Methodology}
\label{sec:method}
\begin{figure}
\centering
\includegraphics[width=0.99\textwidth]{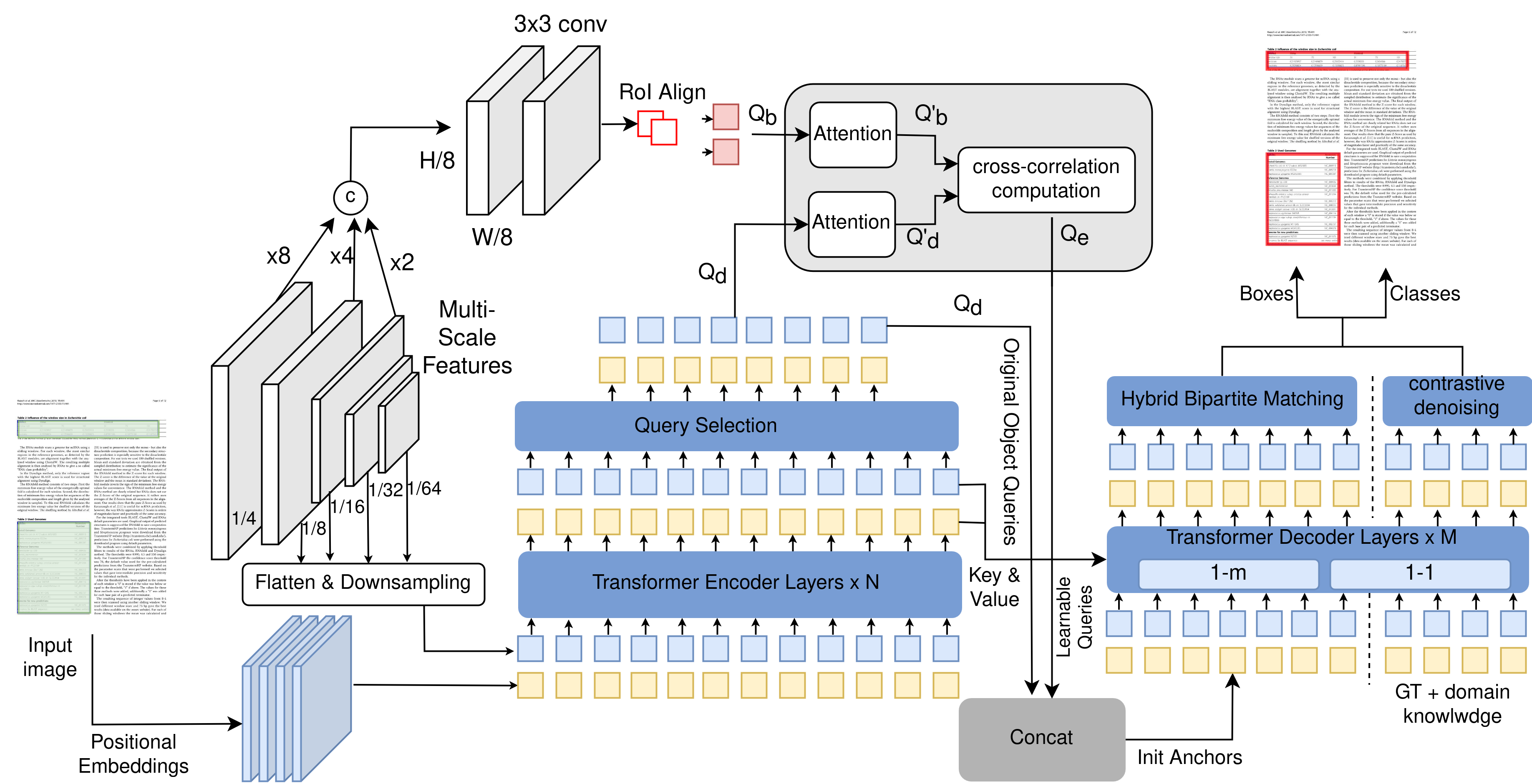}
\caption{Overview of our approach for Document Layout Analysis. The input image is processed through a CNN backbone to extract features, which are then passed to a Transformer encoder-decoder network. The encoder processes the features globally, while the decoder uses object queries to interact with the encoded features and predict bounding boxes and classes for each object in the image. Our approach incorporates an enhanced query encoding mechanism to improve decoder efficiency and a query selection scheme that combines one-to-one and one-to-many matching strategies, improving accuracy and adaptability in identifying various graphical elements across documents.}\label{fig:dino}
\end{figure}

Our approach consists of four integral parts: First, a CNN-based backbone network for extracting multi-scale features from document images. Second, a transformer-based model is employed to detect graphical elements like titles, figures, tables, and text on the pages. Third, we introduce an improved query encoding mechanism, optimizing the model's decoder phase to process complex document layouts more effectively. Fourthly, we implement a unique query selection scheme, blending the decoder's one-to-one matching with a new one-to-many strategy, enhancing accuracy in identifying various graphical elements during training. These modules are collectively trained in an end-to-end manner. 
The complete overview of our approach is shown in Fig.~\ref{fig:dino} and explained in detail in the subsequent subsections.
\subsection{Backbone Multi-scale Features Network}
For processing an input image I of size H×W×3, we use a ResNet-50 backbone network to generate a series of feature maps at reduced resolutions: 1/4, 1/8, 1/16, 1/32, and 1/64 of the original size. Each map is refined using a 1×1 convolution layer, which is crucial for reducing the channel count. This step is essential to control the number of trainable parameters, making the process manageable, especially with limited computational resources. After this reduction, each feature map has 256 channels, which are then input into the transformer network to detect graphical objects on the page.
\subsection{Document Layout Analysis with the transformer Framework}
Recent progress in Transformer-based object detection~\cite{shehzadi_DT_review6,deformable_detr,UPDETR_CVPR20,sparse_semi_detr1,yolos6,fpdetr,rego2,dino23} has revolutionized document analysis. These advanced methods outperform previous models like Faster-RCNN~\cite{faster23} and Mask-RCNN~\cite{mask86}, mainly because they don't require manual techniques such as anchor generation process and NMS. Our approach employs the DINO~\cite{dino23} model, a state-of-the-art network, to detect graphical elements in document images. Our approach includes a transformer network with a unique structure. It has an encoder that processes variously scaled feature maps from a CNN backbone and a decoder that generates the final results. The encoder's task is to generate detailed proposals for graphical elements in the pages, guiding the decoder's positional embeddings for the queries. This decoder uses a deformable attention mechanism for better efficiency in self and cross-attention processes. It also applies contrastive denoising for the object queries, helping the model learn faster. It is highly adaptable, especially during shifts in document types, and it focuses on lower-dimension image features, which often need more data to be included in traditional transformer training. The effectiveness of our transformer-based approach is validated through its impressive performance in detecting graphical page objects on well-known benchmarks like PubLayNet, DocLayNet, and PubTables.

\subsection{Query Encoding Strategy}
\label{sec:Query_Selection}
In the query encoding strategy, we enhance the query mechanism to improve the detection of small graphical objects in document images by combining backbone query features with decoder original queries. This approach creates high-quality object queries, increasing accuracy in identifying the small elements within an image. Here's a detailed explanation:\\

\noindent\textbf{High-level Query Features from Backbone:}~
In our approach, we initially extract high-level features from the early layers of a CNN backbone, such as the ResNet-50. These initial layers are adept at capturing intricate details and textures, including edges, corners, and specific patterns. This level of granularity is crucial for identifying smaller objects within an image. For each processed image, we adjust the dimensions of feature maps \(C4\) and \(C5\) to align with \(C3\) and then concatenate them. The combined feature map undergoes processing through two \(3 \times 3\) convolutional layers, resulting in a feature map \(C_{\text{h}}\) comprising 64 channels. Then, we employ the RoIAlign algorithm~\cite{mask-rcnn84} to extract features based on its bounding box $b_j = (x_{j1}, y_{j1}, x_{j2}, y_{j2})$ with many MLPs , where $(x_{j1}, y_{j1})$ and $(x_{j2}, y_{j2})$ are the coordinates of the upper-left and lower-right corners, respectively. The high-level query features are then defined as:
\begin{equation}
 Q_h = MLP ( \text{RoIAlign}(F_{\text{h}}, b_j)) 
\end{equation}
Here, \(Q_h\) refers to the query features and \(F_h\) denotes features from backbone. Next, we apply a self-attention mechanism, as described in \cite{att75}, to the high-level query features \(Q_h\) and decoder orignal query features \(Q_d\). This mechanism enables the model to prioritize and weight the importance of different aspects of the high-level features, thus enhancing the overall feature representation. Following the self-attention step, we determine the cross-correlation (similarity) between the outputs \(Q'_h\) and \(Q'_d\), using cosine similarity~\cite{cosine_similarity45}. The process is formalized as follows:
\begin{equation}
Q_e = similarity (Q'_d, Q'_h)
\end{equation}
Here, \(Q'_h\) refers to the query features from the backbone after self-attention, while \(Q'_d\) denotes the original decoder queries after self-attention. The final step involves integrating these refined queries \(Q_e\) with the original queries from the decoder, enhancing the overall feature extraction and analysis.\\

\noindent\textbf{Combining Features for Enhanced Detection:} 
In the next step, we integrate two distinct types of features to boost detection capabilities. High-level features, represented by $Q_h$, allow the model to understand the overall layout and context of the document. On the other hand, we have the original transformer query features, which are adept at capturing specific object information. The enhanced features $Q_e$ are obtained from self-attention on $Q_h$ and decoder original queries $Q_d$. This concatenation is particularly beneficial for detecting small graphical elements, such as page headers, footers, and titles, which might be missed by the decoder's original query features alone. Combining these features enhances the model's detection sensitivity to these smaller elements. The combined query features, which we denote as $Q_t$, are formed by concatenating the decoder original query features with the enhanced query features $Q_e$:
\begin{equation}
Q_t = \text{Concat}(Q_d, Q_e)
\end{equation}
This combination of features from both the high-level and the decoder queries enriches the feature set provided to the model, leading to a more robust detection mechanism for various objects within complex documents.\\

\noindent\textbf{Integration with Decoder's Original Queries:}
 By merging previously generated queries with the original decoder queries, our model performs better in identifying elements in document images. This integration enhances the model's ability to detect prominent and subtle features within complex document layouts, making it especially effective for predicting small or easily overlooked objects. The process of query integration and output generation is formulated as:
\begin{equation}
o= Decoder(Q_t, E|A)
\end{equation}
Here, our model utilizes a set of decoder queries, denoted by \( Q_t \), and corresponding outputs from the Transformer decoder, represented as \( o \). The refined image features, processed by the Transformer encoder, are symbolized by \( F \), while \( A \) represents the attention mask, specifically designed for the denoising task~\cite{dn42}.
In this way, this query mechanism combines the strengths of both abstract and detailed image features, facilitating thorough and precise detection of diverse elements within intricate document structures.
\subsection{Query Selection Strategy}
Our research introduces an innovative hybrid matching scheme for analyzing complex documents. This approach uniquely combines two query strategies, one-to-one and one-to-many matching, to enhance the detection and understanding of various elements in detailed documents. Initially, we observe that the one-to-many strategy led to duplicate predictions, as shown in Table~\ref{tab:nms}. To optimize this, we utilized one-to-many matching during the first half of our training iterations, then shifted to one-to-one matching for the remainder. This transition markedly improved accuracy and reduced duplications.

 As a key feature of our hybrid approach, the one-to-many matching branch is designed to enhance object detection in complex document layouts. This innovative branch enables the association of a single detected object with multiple ground truths, a significant advancement over traditional one-to-one matching methods. We integrate original decoder queries with high-quality object queries generated in the one-to-many strategy. These object queries are generated by merging high-level query features from the backbone as explained in subsection~\ref{sec:Query_Selection}. It is particularly useful in complex document layouts where traditional one-to-one matching might struggle. By enabling an object to be matched with several ground truths, the model better understands the document's content, especially in overlapping or closely packed elements. The total loss in ono-to-many strategy is as follows:

\begin{equation}
L^{1-m}_{cls} = \sum_{i=1}^{N_{obj}} |\hat{g}_i - p_i| \cdot BCE(p_i, \hat{g}_i) + \sum_{j=1}^{N_{no}} p_j \cdot BCE(p_j, 0)
\end{equation}

\begin{equation}
L^{1-m}_{reg} = \sum_{i=1}^{N_{obj}} \hat{g}_i \cdot  \mathcal{L}_{GIoU}(bx_i, \hat{bx}_i) + \sum_{i=1}^{N_{obj}} \hat{g}_i \cdot \mathcal{L}_{L1}(bx_i, \hat{bx}_i)
\end{equation}

\begin{equation}
L^{1-m} = L^{1-m}_{cls} + L^{1-m}_{reg}
\end{equation}
where $\hat{g}_i$ is the ground truth, $p_i$ is the actual prediction.
In the one-to-one matching branch, a traditional approach in object detection models, each detected object is directly aligned with a corresponding ground truth. This method is straightforward and effective in scenarios where objects are clearly separated and easily identifiable. It eliminates duplications generated in the one-to-many strategy, ensuring more accurate predictions. The total loss in the one-to-one matching strategy is as follows:
\begin{equation}
L^{1-1} = L^{1-1}_{cls} + L^{1-1}_{reg}
\end{equation}
This hybrid approach retains the benefits of the traditional method, like eliminating the need for Non-Maximum Suppression (NMS), and does not add any extra computational cost. The combination of these two methods in a single model allows for more accurate and efficient object detection in a wide range of scenarios, significantly improving the performance of document analysis tasks.
\section{Experimental Setup}
\label{sec:exp}
\noindent\textbf{Datasets and Evaluation Criteria.}
\label{sec:dataset}
Our study employs three benchmark datasets to evaluate the efficacy of the proposed method: PubLayNet~\cite{PubLayNet3} PubTables~\cite{pubtables5} and DocLayNet~\cite{doclaynet_data}. 
We adopt the mean Average Precision (mAP) metric in line with COCO-style~\cite{coco14} standards to evaluate our approach. We compute precision across a spectrum of Intersection over Union (IoU) thresholds, from 0.50 to 0.95, increasing in 0.05 steps. This IoU range is essential for evaluating our model's accuracy in category-specific tasks. Our mAP calculation, averaged across these IoU levels, follows the established Microsoft COCO benchmark, facilitating a standardized comparison with other models. We further refine our assessment by calculating Average Precision (AP) at specific IoU thresholds of 0.50 and 0.75, offering a focused analysis of the model's performance at these recognized benchmarks. It clearly explains our model's proficiency in accurately classifying various categories.

\noindent\textbf{Implementation Details.}
\label{sec:implement}
Our network is trained on RTXA600 GPUs, utilizing a ResNet-50 network as the backbone, which is pre-trained on ImageNet. We employ the AdamW algorithm for optimization, with a batch size of 16. The training duration is set to 12 epochs for both PubLayNet and PubTables datasets, and extended to 24 epochs for the DocLayNet dataset. We implement a learning rate reduction strategy, decreasing it by a factor of 10 later in the training process. Our approach includes a multi-scale training technique, where images are resized to various lengths without exceeding a maximum size limit. For the testing phase, we resize images to have a shorter side of 640, optimizing image handling during model evaluation.
\begin{table}[ht]
\centering
\caption{\textbf{Evaluation on the DocLayNet Benchmark.} A comparative analysis of outcomes on the DocLayNet Test Dataset. Here, Mask represents Mask R-CNN and Faster indicates Faster R-CNN.
In this comparison, the performances of Mask R-CNN, Faster R-CNN, and YOLOv5 are referenced from~\cite{doclaynet_data}, and the results for the DINO model are derived from~\cite{hybrid_dino98}. The best results are highlighted in bold.}
\begin{tabular}{l|c|c|c|c|c|c}
\hline
\textbf{Classes} & \textbf{Mask} & \textbf{Faster} & \textbf{YOLOv5} & \textbf{DINO} & \textbf{Zhong et al.}~\cite{hybrid_dino98} & \textbf{Ours} \\
\hline
Caption         & 71.5 & 70.1 & 77.7 & 85.5 & 83.2 & \textbf{85.6} \\
Footnote        & 71.8 & 73.7 & \textbf{77.2} & 69.2 & 69.7 & 70.0 \\
Formula         & 63.4 & 63.5 & \textbf{66.2} & 63.8 & 63.4 &  64.7 \\
List-item       & 80.8 & 81.0 & 86.2 & 80.9 & \textbf{88.6} &  83.5 \\
Page-footer     & 59.3 & 58.9 & 61.1 & 54.2 & 90.0 &  \textbf{91.3}\\
Page-header     & 70.0 & 72.0 & 67.9 & 63.7 & 76.3 &  \textbf{77.8} \\
Picture         & 72.7 & 72.0 & 77.1 & 84.1 & 81.6 &  \textbf{84.7}\\
Section-header  & 69.3 & 68.4 & 74.6 & 64.3 & \textbf{83.2} &  82.9\\
Table           & 82.9 & 82.2 & 86.3 & 85.7 & 84.8 &  \textbf{86.1}\\
Text            & \textbf{85.8} & 85.4 & 88.1 & 83.3 & 84.8 &  85.4\\
Title           & 80.4 & 79.9 & 82.7 & 82.8 & 84.9 &  \textbf{86.3}\\
\hline
\textbf{All}    & 73.5 & 73.4 & 76.8 & 74.3 & 81.0 & \textbf{81.6} \\
\hline
\end{tabular}
\label{tab:DocLayNet_comparison}
\end{table}
\section{Results and Discussion}
\label{sec:results}
\subsection{DocLayNet}
Table~\ref{tab:DocLayNet_comparison} summarizes the performance of our approach compared to other approaches on the DocLayNet dataset, with results measured in mean Average Precision (mAP) for different document elements. Our method outperforms previous networks like Mask R-CNN~\cite{mask-rcnn84}, Faster R-CNN~\cite{FasterTD}, YOLOv5~\cite{yolov4}, DINO, and the document analysis approach of Zhong et al.~\cite{hybrid_dino98}, particularly in recognizing 'Caption,' 'Page-footer,' 'Page-header,' and 'Title,' achieving the highest overall mAP at 81.6\%. This comprehensive evaluation across various classes highlights the effectiveness of our approach in accurately detecting and classifying elements in a wide array of document layouts. 
\begin{figure}
\centering
\includegraphics[width=.86\textwidth]{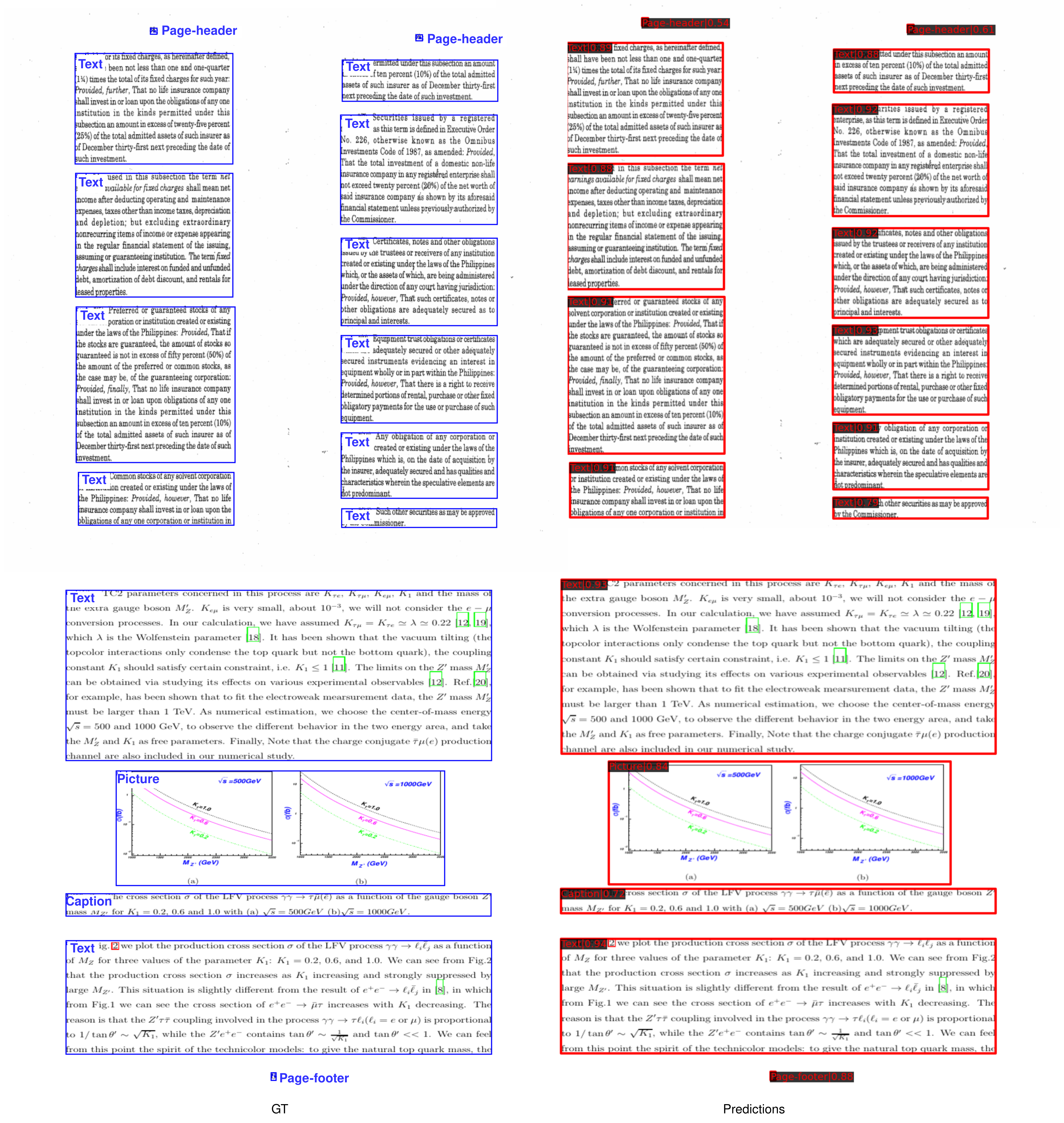}
\caption{Visual analysis of our approach on the DocLayNet dataset. Here, blue color represents ground truth, red denotes prediction by our approach. It illustrates the model's proficiency in identifying small layout elements, specifically highlighting its accuracy in detecting page titles, headers, and footers.}\label{fig:dino_results}
\end{figure}

Fig.~\ref{fig:dino_results} illustrates the visual results of our document layout analysis approach on the DocLayNet dataset. It displays document page with our model's predictions compared to the actual ground truth (GT). In these visual examples, ground truth annotations are outlined in blue, and our model's predictions are in red. This comparison aims to showcase our model's precision in identifying small layout elements, such as page titles, headers, and footers. Using contrasting colors demonstrates the accuracy of our approach in detecting and classifying the intricate details of document layouts.

\subsection{PubLayNet}
We also evaluate and compare our approach with previous document analysis approaches on the PubLayNet dataset. The results of these comparisons are detailed in Table~\ref{tab:publaynet_comparison}. 
\begin{table}[ht]
\centering
\caption{\textbf{Evaluation on the PubLayNet Benchmark.} A comparative analysis of results on the PubLayNet Validation Set. The results highlight the effectiveness of our approach. The best results are highlighted in bold.}
\label{tab:publaynet_comparison}
\renewcommand{\arraystretch}{1}
\begin{tabular}{l|c|c|c|c|c|c}
\hline
\textbf{Method}  & \textbf{Text} & \textbf{Title} & \textbf{List} & \textbf{Table} & \textbf{Figure} & \textbf{mAP} \\ 
\hline
Faster R-CNN~\cite{PubLayNet3} & 91.0 & 82.6 & 88.3 & 95.4 & 93.7 & 90.2 \\ 
Mask R-CNN~\cite{PubLayNet3}  & 91.6 & 84.0 & 88.6 & 96.0 & 94.9 & 91.0 \\ 
Naik et al.~\cite{naik86}  & 94.3 & 88.7 & 94.3 & 97.6 & 96.1 & 94.2 \\ 
Minouei et al.~\cite{DLA_IPRIA21} & 94.4 & 90.8 & 94.0 & 97.4 & 96.6 & 94.6 \\ 
DiT-L~\cite{Lidit78} & 94.4 & 89.3 & 96.0 & 97.8 & 97.2 & 94.9 \\
SRRV~\cite{SRRV34} & 95.8 & 90.1 & 95.0 & 97.6 & 96.7 & 95.0 \\
DINO~\cite{dino23} & 94.9 & 91.4 & 96.0 & 98.0 & 97.3 & 95.5 \\
TRDLU~\cite{trans_DLU5} & 95.8 & 92.1 & \textbf{97.6} & 97.6 & 96.6 & 96.0 \\
UDoc~\cite{unidoc89} & 93.9 & 88.5 & 93.7 & 97.3 & 96.4 & 93.9 \\
LayoutLMv3~\cite{layoutMV3} & 94.5 & 90.6 & 95.5 & 97.9 & 97.0 & 95.1 \\
VSR~\cite{vsr45} & 96.7 & 93.1 & 94.7 & 97.4 & 96.4 & 95.7 \\
Zhong et al.~\cite{hybrid_dino98}  & 97.4 & 93.5 & 96.4 & 98.2 & 97.2 & 96.5 \\
\hline
Our  & \textbf{98.0} & \textbf{94.2} & 97.3 & \textbf{98.6} & \textbf{98.5} & \textbf{97.3}  \\
\hline
\end{tabular}
\end{table}
The results indicate that our approach significantly outperforms previous methods, demonstrating its superior performance in document analysis.
\subsection{PubTables}
We also evaluate our approach and compare it with previous table detection approaches on PubTables dataset. The results of these comparisons are detailed in Table~\ref{tab:PubTables}. 
The results clearly demonstrate that our approach outperforms previous table detection approaches, highlighting its effectiveness and efficiency in accurately identifying and classifying table elements within complex documents.
\begin{table*}[ht]
\centering
\caption{Comparative Analysis of Results on the PubTables Validation Set. The best results are highlighted in bold.}
\label{tab:PubTables}
\renewcommand{\arraystretch}{1}
\begin{tabular*}{.7\textwidth}
{@{\extracolsep{\fill}}ccccc@{\extracolsep{\fill}}}
\toprule
\textbf{Method} & 
\textbf{Detector} & 
\textbf{mAP} & 
\textbf{AP\textsuperscript{50}} &
\textbf{AP\textsuperscript{75}}   \\
\toprule
Smock et al.~\cite{pubtables5} & Faster R-CNN   & 82.5 & 98.5 & 92.7  \\
Smock et al.~\cite{pubtables5}  & DETR  & 96.6 & 995 & 98.8 \\ 
Minouei et al.~\cite{TDcl34} & Sparse R-CNN+PVT  & 98.2 & - & - \\
Our & DINO  & \textbf{98.6} & \textbf{99.8} & \textbf{99.1}    \\
\bottomrule
\end{tabular*}
\end{table*}
\subsection{Ablation Study}
In our ablation study, we explore the impact of object query selection, the effectiveness of matching strategies, and the influence of the quantity of learnable queries in our Transformer-based model. This investigation is designed to observe how these key components individually and collectively affect our model's precision and functionality in analyzing complex document layouts.

\begin{table}[ht]
\centering
\caption{Detailed Ablation Analysis on the PubLayNet Validation Dataset.}
\renewcommand{\arraystretch}{1.5}
\begin{tabular*}{0.95\textwidth}
{@{\extracolsep{\fill}}ccccccc@{\extracolsep{\fill}}}
\hline
\textbf{Method} & \textbf{Text} & \textbf{Title} & \textbf{List} & \textbf{Table} & \textbf{Figure} & \textbf{mAP} \\
\midrule
DINO-Queries ($Q_d$) & 94.9 & 91.4 & 96.0 & 98.0 & 97.3 & 95.52 \\

Hybrid-Queries ($Q_d + Q_e$) &  98.0 & 94.2 & 97.3 & 98.6 & 98.5 & 97.3 \\
\hline
\end{tabular*}
\label{tab:publaynet_scores}
\end{table}

\noindent\textbf{Influence of object query selection}
In the ablation study, we observe the impact of object query selection, which is crucial for detecting small graphical objects like page headers, footers, and titles in document layout analysis. The study examines the enhanced query mechanism that combines high-level backbone features with decoder original query features. By integrating the refined query features with the original queries, we observe a significant improvement, as shown in Table~\ref{tab:publaynet_scores}, to accurately predict and identify smaller elements within document layouts. This comparison shows how high-quality object queries improve document analysis. It demonstrates that modifying query integration can significantly enhance the model's performance and accuracy.
\begin{table}[ht]
\centering  
\caption{Performance comparison using various query combinations as input to the decoder on PubLayNet dataset. Here, $Q_d$ represents the original decoder queries, while $Q_e$ signifies the enhanced queries. The highest mean Average Precision (mAP) is achieved by combining the original DINO queries with the enhanced queries, indicating improved performance during training with overlap in predictions. A training approach that uses $Q_d + Q_e$ for the initial half of training epochs before switching to $Q_d$ is shown to be effective.} \label{tab:nms}
\begin{tabular*}{.6\textwidth}
{@{\extracolsep{\fill}}cccccc@{\extracolsep{\fill}}}
\toprule
$\mathbf{Q_d}$ &
$\mathbf{Q_d + Q_e}$ &
\textbf{NMS-free} & 
\textbf{mAP} & 
\textbf{AP\textsuperscript{50}} &
\textbf{AP\textsuperscript{75}}   \\
\toprule
\checkmark  & \color{black}\xmark  & \checkmark  & 95.5 & - & -  \\
\hline
\color{black}\xmark   & \checkmark  & \color{black}\xmark  & 98.4 & 98.8  & 97.7   \\
\hline
\checkmark & \checkmark  & \checkmark  & 97.3 & 98.5 & 97.4   \\
\midrule
\end{tabular*}
\vspace{-1em} 
\end{table}

\noindent\textbf{Influence of matching strategy}
In our document layout analysis approach, employing one-to-one and one-to-many matching strategies provides a comprehensive approach, as shown in Table~\ref{tab:nms}. In our setup, $Q_d$ represents the standard decoder queries, while $Q_e$ represents the enhanced queries. The data indicates that the best mean Average Precision (mAP) is obtained when these two sets of queries are used together, which leads to better training results. It's particularly effective to start training with $Q_d$ and $Q_e$ and then transition to using just $Q_d$ halfway through the training process. One-to-one matching using $Q_d$, aligning each prediction with a single ground truth, is efficient for clear, distinct objects, ensuring straightforward training. On the other hand, one-to-many matching employing $Q_d + Q_e$ allows a single prediction to correspond to several ground truths, adeptly handling complex layouts with overlapping or closely packed elements. This dual strategy leverages the strengths of both approaches, enhancing the model's ability to accurately detect and classify a wide range of object types in various document layouts.

\noindent\textbf{Influence of Learnable queries Quantity}
The quantity of learnable queries in Transformer-based models like DINO significantly affects their performance in document layout analysis, as observed in Table~\ref{tab:queries}. More queries enable finer detection of detailed elements and improve overall accuracy, but there's a need for balance. Excessive queries can increase computational demands and risk overfitting, while too few may miss intricate details. Thus, optimizing the number of queries is crucial for efficient processing, balancing computational resources, and ensuring adaptability across various document types and complexities.
\begin{table}[ht]
\begin{center}
\caption{Performance comparison using different numbers of learnable queries to the decoder input on PubLayNet Dataset. The best-performing results are highlighted in bold, illustrating the optimal number of queries required for best model performance. As indicated, the model generally improves with more queries, up to a point, after which the performance decreases, suggesting an optimal query range for efficient detection across various object sizes.}
\label{tab:queries}
\renewcommand{\arraystretch}{1} % Default value: 1
\begin{tabular*}{.6\textwidth}
{@{\extracolsep{\fill}}ccccccc@{\extracolsep{\fill}}}
\toprule
\textbf{N} & 
\textbf{AP} & 
\textbf{AP\textsuperscript{50}} &
\textbf{AP\textsuperscript{75}} & 
\textbf{AP\textsubscript{s}} & 
\textbf{AP\textsubscript{m}} &
\textbf{AP\textsubscript{l}}    \\
\toprule
 100 & 95.3 & 96.7 & 95.8 & 35.8 & 65.3 & 89.2    \\
 
 200 & 96.5 & 97.3 & 96.6 & 43.5 & 71.8 & 96.4   \\
\textbf{300} & \textbf{97.3} & \textbf{98.5} & \textbf{97.4} & \textbf{43.8} & \textbf{72.7} & \textbf{96.7}  \\
 400  & 96.4  & 98.2 & 97.0 & 43.1 & 60.7 & 96.1\\
\bottomrule
\end{tabular*}
\end{center}
\end{table}
\section{Conclusion}
\label{sec:conclusion}
This paper introduces a approach for analyzing document layouts, focusing on accurately identifying elements like text, images, tables, and headings in documents. We introduce a hybrid query mechanism that enhances object queries for contrastive learning, improving the efficiency of the decoder phase in the model. Moreover, during training, our approach features a hybrid matching scheme that combines the decoder's original one-to-one matching with a one-to-many matching branch, aiming to increase the model's accuracy and flexibility in detecting diverse graphical elements on a page. We evaluate our approach on benchmark datasets like PubLayNet, DocLayNet, and PubTables. It demonstrates superior accuracy and precision in layout analysis, outperforming current state-of-the-art methods. These advancements significantly aid in transforming document images into editable and accessible formats, streamlining information retrieval and data extraction processes. The implications of our research are substantial, affecting areas such as digital archiving, automated form processing, and content management systems. This work represents a significant contribution to document analysis and digital information management, setting new benchmarks and paving the way for future advancements.

\bibliographystyle{IEEEtran}
\bibliography{main}% common bib file

\end{document}